 \documentclass[pmlr,twocolumn,10pt]{jmlr} 

\usepackage{booktabs}
\usepackage{afterpage}
\usepackage{caption}
\usepackage{authblk} 

\theorembodyfont{\upshape}
\theoremheaderfont{\scshape}
\theorempostheader{:}
\theoremsep{\newline}

\title{Applying Large Language Models for Causal Structure Learning in Non Small Cell Lung Cancer}

\author{
\Name{Narmada Naik}\textsuperscript{1},
\Name{Ayush Khandelwal}\textsuperscript{1},
\Name{Mohit Joshi}\textsuperscript{1},
\Name{Madhusudan Atre}\textsuperscript{1},
\Name{Hollis Wright}\textsuperscript{1},
\Name{Kavya Kannan}\textsuperscript{1},
\Name{Scott Hill}\textsuperscript{1},
\Name{Giridhar Mamidipudi}\textsuperscript{1},
\Name{Ganapati Srinivasa}\textsuperscript{1},
\Name{Carlo Bifulco}\textsuperscript{2},
\Name{Brian Piening}\textsuperscript{2},
\Name{Kevin Matlock}\textsuperscript{1}
}
\affil{{\textsuperscript{1}dātma Health Science, Beaverton, OR, USA},
{\textsuperscript{2}Earle A. Chiles Research Institute, Providence Cancer Institute, Portland, OR, USA}}

\begin{document}

\maketitle

\begin{abstract}
Causal discovery is becoming a key part in medical AI research. These methods can enhance healthcare by identifying causal links between biomarkers, demographics, treatments and outcomes. They can aid medical professionals in choosing more impactful treatments and strategies. In parallel, Large Language Models (LLMs) have shown great potential in identifying patterns and generating insights from text data. In this paper we investigate applying LLMs to the problem of determining the directionality of edges in causal discovery. Specifically, we test our approach on a deidentified set of Non Small Cell Lung Cancer(NSCLC) patients that have both electronic health record and genomic panel data. Graphs are validated using Bayesian Dirichlet estimators using tabular data. Our result shows that LLMs can accurately predict the directionality of edges in causal graphs, outperforming existing state-of-the-art methods. These findings suggests that LLMs can play a significant role in advancing causal discovery and help us better understand complex systems.
\end{abstract}
\begin{keywords}
Large Language Models,
Causal Discovery,
Electronic Health Record,
Genomics
\end{keywords}

\section{Introduction}
\label{sec:intro}

Healthcare data analysis has been revolutionized in recent years with the application of Machine Learning (ML) and Deep Learning (DL) techniques \cite{yang_intelligent_2021}. But despite the success in predictive modelling, there is great interest in providing explainable models for the causal relationships between variables \cite{shi_learning_2022}. The current ``black box'' approach to modelling has limited interpretability and has not achieved acceptance in clinical settings \cite{chaddad_survey_2023,linardatos_explainable_2021}. Causal modelling can provide an understanding of the underlying cause-effect relationship of the data, allowing counterfactual analysis to be performed \cite{10.1093/biomet/82.4.669}. There are many situations in which a cause-effect relationship is not known; in such scenarios we must rely on Causal Discovery algorithms to identify causal relationships between variables. State of the art Causal Discovery algorithms rely on score-based or constraint based methods to generate Directed Acyclic Graphs (DAGs) from tabular data sets but they don't incorporate any domain expertise or expert knowledge. 

The recent advancements in Large Language Models (LLMs) have sparked interest in their application for Causal Discovery or Causal Structure Learning (CSL) \cite{zhang2023understanding,liu_magic_2023}. LLMs have been proposed to serve as a surrogate for expert knowledge, domain experience, or similar studies in the field to estimate priors for Causal Discovery. In this article, we focus on incorporating the feedback of LLM for CSL in the medical domain, specifically the field of oncology \cite{SHAPIRO2021271.e1}. We show the potential of applying LLMs in Causal Discovery by improving the accuracy of the generated Causal DAGs on data extracted from the Electronic Health Record (EHR) and Molecular Genomic Reports. Models are scored using the Bayesian Dirichlet (likelihood) equivalent uniform (Bdeu) score. 

We focus our efforts on Non Small Cell Lung Cancer (NSCLC) which accounts for about 85\% of lung cancer cases, while Small Cell Lung Cancer (SCLC) accounts for the remaining 15\% \cite{zappa2016nonsmall}. In oncology molecular tests are already being used in the diagnosis to determine the course of treatment\cite{PIRKER2020283}. But there is a growing interest in finding new causative biomarkers and individualize treatments based on a patients medical history. Causal Modelling has been proposed as a key component in biomarker discovery \cite{cai_systematic_2019}. 


\section{Study Sample (Dataset)}

We start off with a dataset of $>1000$ deidentified patients extracted from Providence St. Joseph Health (PSJH’s) clinical data warehouse. We reduce the cohort to patients that have been diagnosed for NSCLC, bringing the sample size to a total of 455 patients. We only include patients that have a recorded smoking status, leaving a total of 326 patients remaining. The features used for this experiment consist of a multi-modal dataset comprised of Electronic Health Record (EHR), including age, smoking status, biological sex, a variety of presenting symptoms, and cancer stage. We also include somatic mutation status of the genes KRAS, EGFR, FGFR1, ALK, MET, PIK3CA, BRAF, ROS1 and RET. From this set of genes only EGFR, ALK, ROS1, BRAF, NTRK, HER2, MET, RET, KRAS have an FDA approved therapy whereas PIK3CA, AKT1, PTEN are under clinical trials \cite{tan_targeting_2020, sirhan_therapeutic_2023}. The demographics of this data set is shown in Table \ref{tab:demographics}. From this table we see that majority of the cases that have treatment recorded are Stage IV cancers and most patients are treated using standard chemotherapy. In addition, the majority of patients are non-smokers.  Figure \ref{fig:KaplanMeierCurve} shows Kaplan-Meier \cite{goel2010understanding} curves for the treatments given to the patients inside the cohort. The curve labelled ``All Patients'' illustrated the mean survival times of the entire patient cohort. The survival probability for chemotherapy appear to decline over time at a slower rate than the general population.


\begin{table}[htbp]
\caption{Summary of Clinical and Demographic Features}
\label{tab:demographics}
\begin{tabular}{lc}
\toprule
            Characteristic &       Summary \\
\midrule
        Number of Patients &           326 \\
                       Age &     73.3±10.6 \\
             Survival Days & 1179.7±1581.1 \\
                       Sex &               \\
                \quad Male &        42.3\% \\
              \quad Female &        57.7\% \\
            Smoking Status &               \\
              \quad Smoker &        19.0\% \\
          \quad Non-Smoker &        81.0\% \\
                     Stage &               \\
                   \quad I &        30.4\% \\
                  \quad II &         8.0\% \\
                 \quad III &        14.1\% \\
                  \quad IV &        47.5\% \\
            Treatment Plan &               \\
\quad Unknown/Not Recorded &        69.3\% \\
        \quad Chemotherapy &        20.9\% \\
   \quad Targeted Therapy &         7.1\% \\
   \quad Immunotherapy &            2.8\% \\
                  Genomics &               \\
                \quad KRAS &        27.9\% \\
                \quad EGFR &        39.0\% \\
               \quad FGFR1 &         5.8\% \\
                 \quad ALK &        20.6\% \\
                 \quad MET &        11.0\% \\
              \quad PIK3CA &        26.4\% \\
                \quad BRAF &         5.2\% \\
                 \quad RET &        42.9\% \\
\bottomrule
\end{tabular}

\end{table}

\begin{figure}[htp]
\floatconts
  {fig:KaplanMeierCurve}
  {\caption{Kaplan–Meier Survival Curve For Associated Treatment Plans}}
  {\includegraphics[height=5.3cm,keepaspectratio]{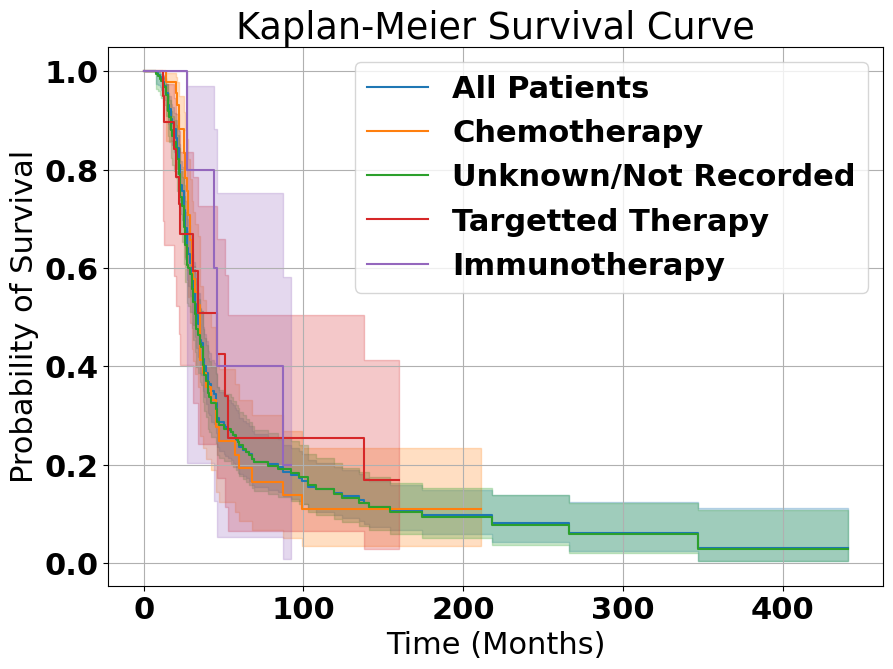}}
\end{figure}

\begin{figure}[htbp]
\floatconts
  {fig:correlation_heatmap}
  {\caption{Heatmap Showing Correlation of Features in the Dataset}}
  {\includegraphics[height=5.7cm,keepaspectratio]{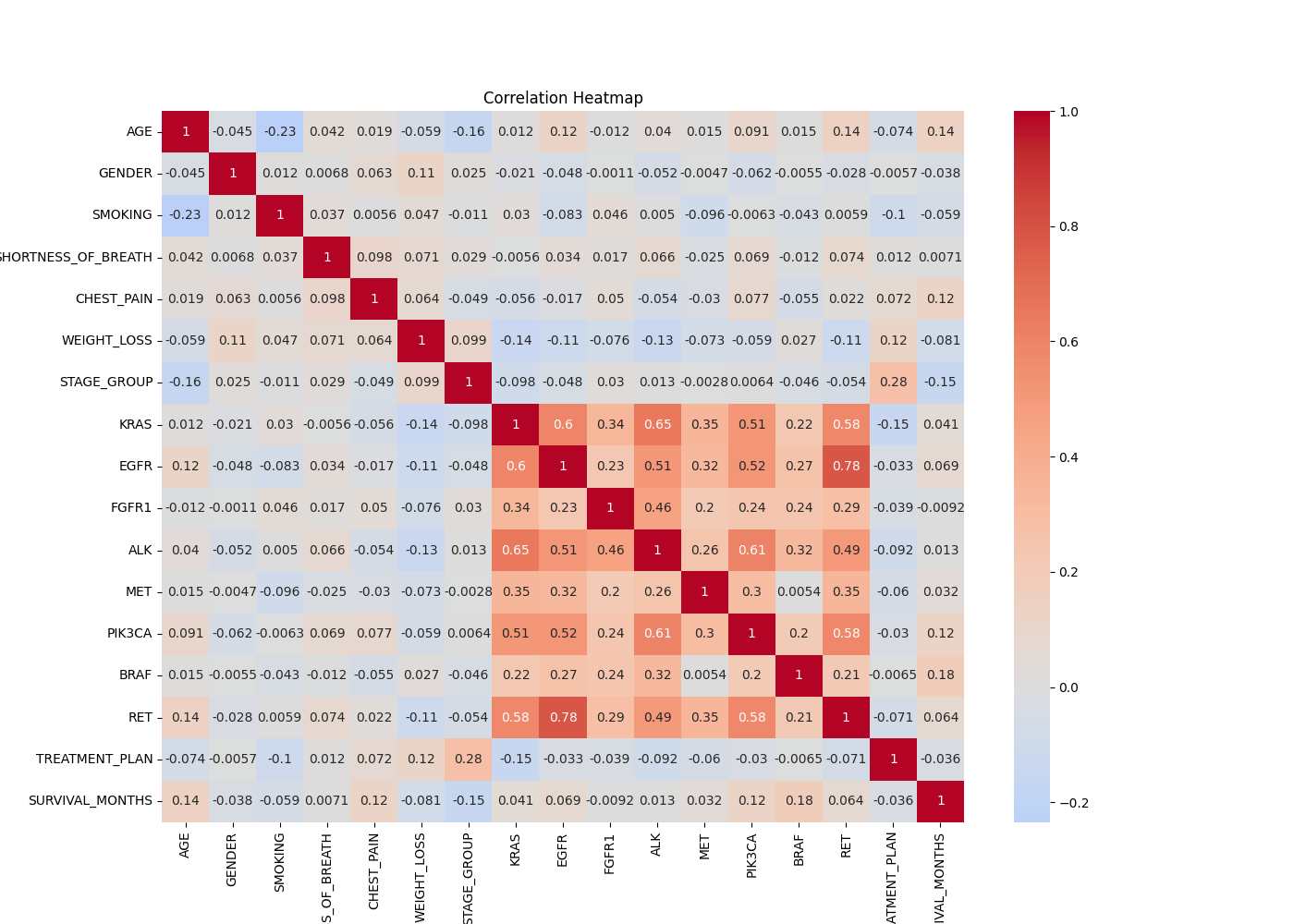}}
\end{figure}

A heatmap showing the correlation between the chosen features is given in Figure \ref{fig:correlation_heatmap}. The correlation coefficient between any two variables is displayed in each cell. The hue of the cell indicates the direction and strength of the correlation: When the correlation is near to 1 , there is a strong positive correlation (the tendency for both variables to rise as one rises). If the correlation is near to -1, it is strongly negative (when one variable rises, the other tends to fall). Little or no association between the variables is indicated by a correlation that is near to 0. From this heatmap we see little correlation amongst the clinical variables but strong correlation among the genomic markers. Despite these correlation, few if any of these genomic markers are causally related. For example, both EGFR and KRAS are strongly associated with smokers in lung cancer. The more likely explanation is that both variables are being connected through a shared parent node.\cite{takamochi_differences_2013}.

\section{Methods}
In this section we present our overall methods for DAG generation, selection of DAG and ATE for different treatment categories. This study was conducted under a protocol approved by the Providence IRB (Protocol 2018000188). 

\subsection{Causal Network Generation}
While discovering causal structures from observational data is a difficult task, a number of strategies have been proposed. Most state of the art strategies rely on optimization techniques such as those used in the PC algorithm \cite{spirtes_causation_1993} and NOTEARS \cite{zheng_dags_2018}. 

Instead we opt to start building the graph using LLMs to form the edges between the nodes. In this scenario the LLM is being used as a surrogate for a domain expert to condition on the DAG. Initial experiments involved using zeroshot prompt \cite{suzgun2022prompt}  engineering to question the existing an edge between each of the 18 feature nodes, requiring a total of $n(n-1)/2= 153$ prompts. Each prompt was formatted to query a potential cause-effect relationship. Afterwards the completions from the LLM are interpreted to determine if there is an edge in the causal graph in the specified direction. Table \ref{tab:prompts} shows a set of example prompt/completion pairs along with the causal edge being tested and final assumption. 

\begin{table*}[htbp]
\floatconts
  {tab:prompts}%
  {\caption{From left to right. An example prompt used to test for the provide edge, the corresponding completion and the final verdict.}}%
  {
\begin{tabular}{|p{3cm}|c|p{7cm}|c|}
\hline
\centering{Prompt} & Edges & \centering{Completion/Response} & Assumption \\
\hline 
Does age effect Gender in NSCLC & Age$\rightarrow$Gender & Age and gender can both independently influence the development and characteristics of non-small cell lung cancer (NSCLC), but they do not have a direct cause-and-effect relationship with each other & No \\
\hline 
Does age effect survival analysis for NSCLC & Age$\rightarrow$Survival & Yes, age can have an impact on survival analysis & Yes \\
\hline Does age causes shortness of breath in NSCLC & Age$\rightarrow$Shortness of breath & Age itself does not directly cause shortness of breath in NSCLC & No \\
\hline Does KRAS mutation effect survival in NSCLC & KRAS$\rightarrow$Survival & KRAS mutation subtype: There are different subtypes of KRAS mutations and some studies have suggested that specific subtypes may have varying impacts on survival & Yes \\
\hline Does targeted therapy effect survival in NSCLC & Treatment Plan$\rightarrow$Survival &  Therapy can have a significant impact on the survival outcomes of patients with NSCLC, particularly those with specific molecular alterations that are targeted by the therapy. & Yes \\
\hline
\end{tabular}
}
\end{table*}

Next, we use a single prompt that contains all the nodes and prompt the LLM to generate the full graph in a single step. Since this method can be done in a single step, it is significantly more computationally efficient. The full prompt and output is shown in Table \ref{tab:operatornames}. 

Afterwards, we started updating the model using human intervention. We first corrected the lack of Age influencing smoking status of a patient with the following prompt, ``how age is not cause smoking please relook into the adjacency matrix and generate a correct one.'' Then we add in that staging and smoking should be effecting the mutation status, ``the stage group and smoking should cause some mutation in nsclc.'' This represents the \emph{V3} LLM DAG. 

For the final model we request the LLM to reinvestigate how mutation is effecting the treatment plan and survival months. Finally, we request the LLM to update the model such that treatment plan should effect survival months. This final adjacency matrix represents the \emph{V5} version of the LLM model.


\begin{table*}[htbp]
    \floatconts
    {tab:operatornames}%
    {\caption{Single Prompt Input and Response}}%
    {
    \begin{tabular}{|p{2cm}|p{10cm}|}
    \hline
    \centering{Prompt} & Generate me a cause effect adjacency matrix for these nodes AGE, SMOKING, GENDER, SHORTNESSOFBREATH, CHESTPAIN, WEIGHTLOSS, TREATMENTPLAN, SURVIVALMONTHS, STAGEGROUP, KRAS, EGFR, FGFR1, ALK, MET, PIK3CA, BRAF, ROS1, RET mutation doesn't cause symptoms. \\
    \hline
    \centering{Response} & In  order to create a cause-effect adjacency matrix, we need to understand the relationships between the given nodes. Here's a possible interpretation of the relationships between them:
    AGE can affect the TREATMENTPLAN and SURVIVALMONTHS. SMOKING can lead to CHESTPAIN, SHORTNESSOFBREATH, and can affect the TREATMENTPLAN, SURVIVALMONTHS, and STAGEGROUP. GENDER can affect the TREATMENTPLAN and SURVIVALMONTHS. SHORTNESSOFBREATH and CHESTPAIN can indicate the STAGEGROUP, which in turn influences the TREATMENTPLAN and SURVIVALMONTHS. WEIGHTLOSS can also indicate the STAGEGROUP and can affect the TREATMENTPLAN and SURVIVAL\_MONTHS. Mutations (KRAS, EGFR, FGFR1, ALK, MET, PIK3CA, BRAF, ROS1, RET) do not cause symptoms (as per the user's instructions) but they can affect the TREATMENTPLAN, SURVIVALMONTHS, and STAGEGROUP. \\
    \hline
    \end{tabular}
    }
\end{table*}

\subsection{Graph Validation}

A persistent problem when applying Causal Discovery in real world scenarios is that it is difficult to validate the generated graph. In addition due to the randomized nature of LLMs, even with low temperature settings response can vary. Both of these scenarios require some metric to determine which graph best fits the available data. Each of these DAGs represent a possible set of causal relationships between variables. Selecting the best or the optimal DAG among these becomes crucial in ensuring the reliable predictions and interpretations. Here we discuss different strategies to tackle this challenge. 





\subsubsection{Bayesian Estimation}

In order to compare the ``fit'' of our proposed network we chose the Bayesian Dirichlet equivalent uniform ($Bdeu$) score \cite{heckerman_learning_1995}. The $Bdeu$ score is an extension of the Maximum Likelihood Probability that includes the prior probability of a given DAG and the marginal probability of the observed data. Mathematically it is written as the following:

\begin{equation}
    Bdeu = log P(D|G) + log P(G) - log P(D)
\end{equation}

For each node a separate $Bdeu$ score is calculated that best matches the Conditional Probability Distribution (CPD) of the parent nodes. When evaluating a singular node with a total number of $N_j$ possible states and parent nodes with a total of $N_i$ possible configurations the $Bdeu$ score simplifies to the following equation:

\begin{equation}
Bdeu = \sum_{i=1}^{N_i} \sum_{j=1}^{N_j} \left(n_{ij} + \frac{\alpha}{N_j}\right) \log\left(\frac{n_{ij} + \frac{\alpha}{N_j}}{n_i + \alpha}\right)
\end{equation}

where $n_ij$ is the number of times the node takes on its $j$-th class given that the parent nodes are in their $i$-th configuration.
$n_i$ is the total number of times the input nodes are in their $i$-th configuration and $\alpha$ is a heuristic parameters which is often used to represent the Equivalent Sample Size (ESS). Intuitively the $Bdeu$ score is often chosen due it's ability to penalize complex models, particular models with large numbers of parent nodes \cite{liu_empirical_2012}. At the same time it does not require prior domain knowledge and is computationally efficient \cite{scutari_dirichlet_2018}. The total $Bdeu$ score of a model is simply the sum of all $Bdeu$ scores for each node.

\begin{figure*}[!ht]
\floatconts
  {fig:Causal DAG LLM sequence}
  {\caption{Causal DAG LLM sequence prompts and adjacency matrix}}
  {\includegraphics[height=4cm,keepaspectratio]{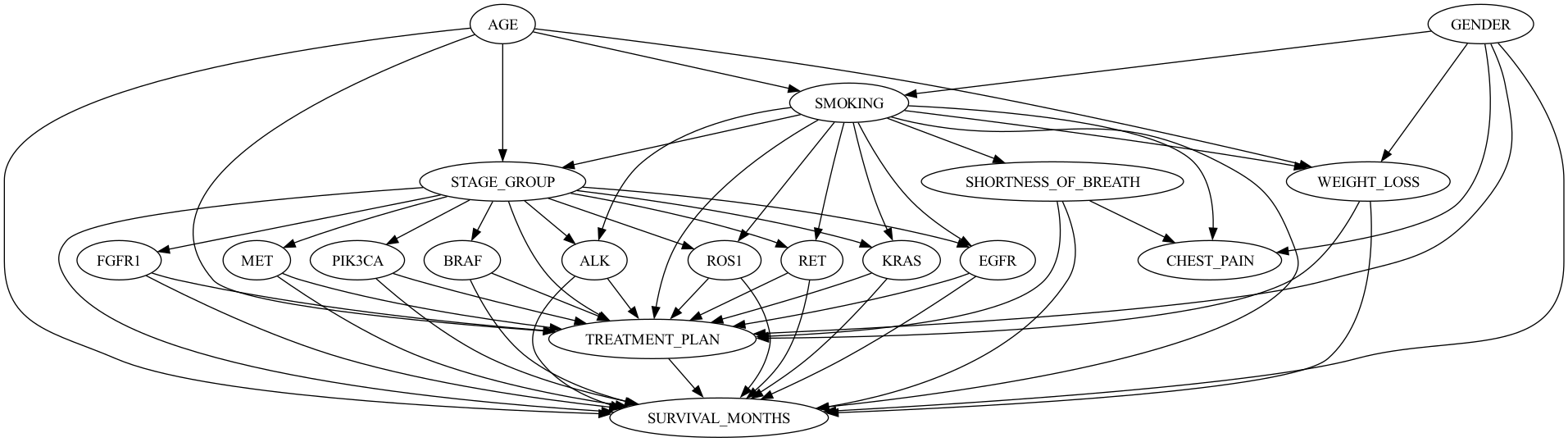}}
\end{figure*}

\begin{figure*}[htbp]
\floatconts
  {fig:llm_graph_codeinterpreter_v3}
  {\caption{DAG Generated for the \emph{V3} version of LLM Model}}
  {\includegraphics[height=2.5cm,keepaspectratio]{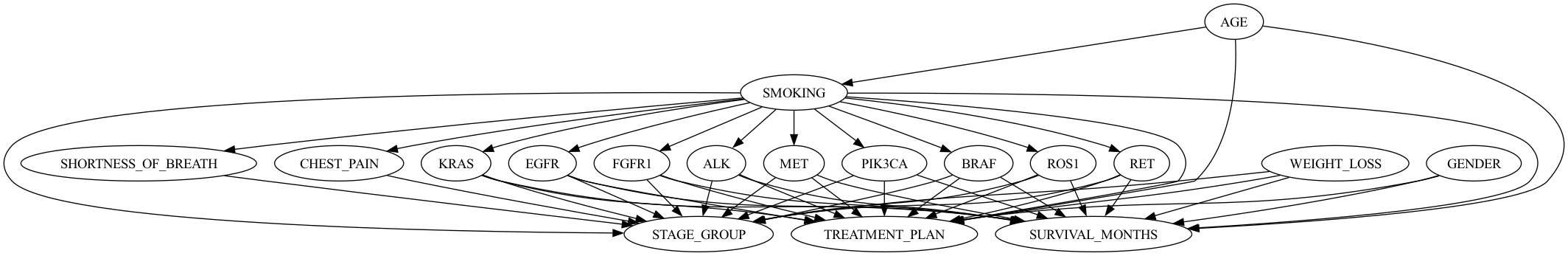}}
\end{figure*}

\section {Interpretability and Results}
A key advantage of using causal modelling is that they ensure trust in research and healthcare policies, emphasise the effects of interventions and make sure decisions are based on accurate evidence \cite{almeda_causal_2019}. In this section we will discuss the DAGs that can be generated using the the LLM responses. The main method used for interpreting and validation is by modelling these DAGs as Bayesian networks. Bayesian networks are natural models for Causal DAGs, and have been shown to effective at inferring and evaluating outcomes of causal inference \cite{pearl_bayesian_1995}. Fitting and evaluating the Conditional Probability Distributions for each node in the network is done using the Bdeu score \cite{liu_empirical_2012}. For measuring the intervention effect, we calculate the Average Treatment Effect (ATE) given both mutation status and treatment combination.

\subsection{ DAG Generation }

The first DAG is created using the iterative prompt approach utilizing OpenAI's GPT-4 model to process the prompts \cite{openai_gpt-4_2023}. The output of the model is manually processed and transformed into an adjacency matrix which is then displayed as a DAG. The DAG for the iterative method is shown in Figure \ref{fig:Causal DAG LLM sequence}. From this Figure we see that this method assigns all nodes to directly effecting the survival of a patient with the exception of the symptom chest pain. It should also be noted that it puts the main cause of mutations to be the cancer stage in combination with the smoking status of individuals. 

For the single prompt method, we utilized OpenAI's Codeinterpreter plugin for GPT-4 that parses the input into a singlular edge adjacency matrix. The DAG for this method is shown in Figure \ref{fig:llm_graph_codeinterpreter_v3}. Unlike in Figure \ref{fig:Causal DAG LLM sequence}, we see significantly less nodes directly interacting with patient's survival time. In addition mutations are not only influenced by a patient's smoking status, not their tumor stage. However the most glaring omission is the fact that neither stage nor treatment effects a patients overall survival.

Our final method of integrating the single prompt with additional corrections is given in Figure \ref{fig:llm_graph_codeinterpreter_v5}. Main benefits of the provided corrections are that treatment plan and survival directly cause the outcome of a patient. 

\begin{figure*}[htbp]
\floatconts
  {fig:llm_graph_codeinterpreter_v5}
  {\caption{DAG For the \emph{V5} LLM Model.}}
  {\includegraphics[height=4cm,keepaspectratio]{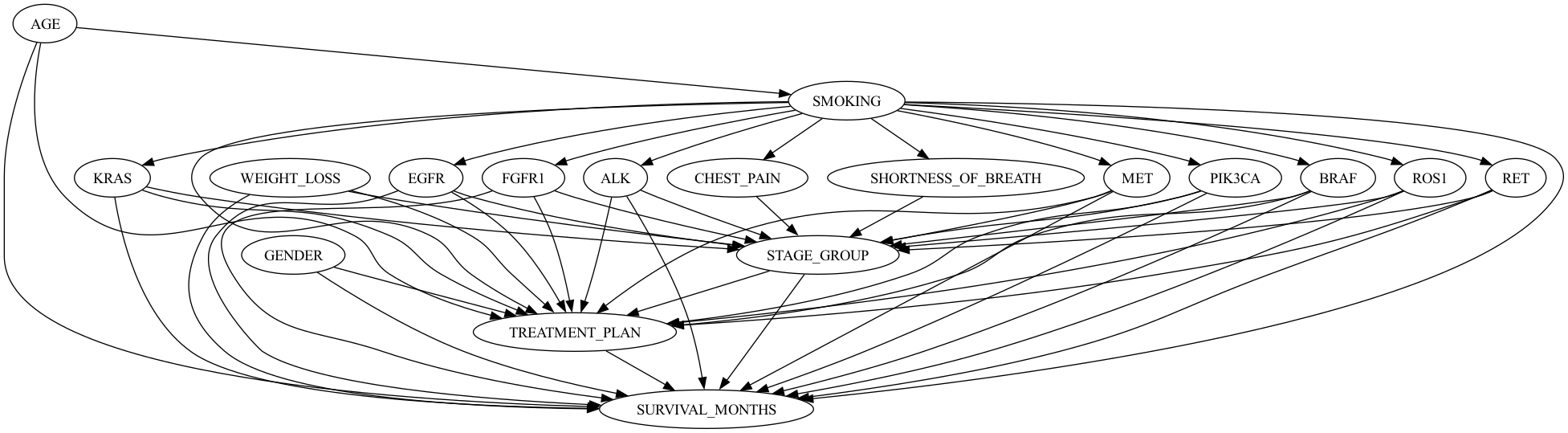}}
\end{figure*}

To compare our methods to existing causal discover algorithms, we applied the NOTEARS and PC algorithms to our data set. The NOTEARS algorithm learns a graph structure by minimizing a continuous, differentiable objective function \cite{zheng_dags_2018}. While optimizing the graph, it NOTEARS includes an ``acyclic constraint'' to ensure the final graph is a true DAG. The PC algorithm instead starts with a fully connected graph and performs statistical independence tests to removes edges. After removing non-statistically dependence nodes, the algorithm enforces acyclicity on the remaining edges using a set of four rules \cite{zhang_completeness_2008}. Most importantly, neither algorithm takes any prior-knowledges while generating their graphs, relying on observed data alone. Thee NOTEARS implementation was done using the \emph{causalnex} library while the PC algorithm used the \emph{gcastle} library \cite{Beaumont_CausalNex_2021,zhang_gcastle_2021}. Figure \ref{fig:no_tears} shows the DAG generated based on NOTEARS algorithm and Figure \ref{fig:PC} shows DAG with PC algorithm. It should be noted that both methods are unable to generate clinical relevance DAGs using the provided dataset. For example, in the  NOTEARS algorithm DAG we have several symptoms causing AGE in addition to a view somatic mutations. The generated DAGs based on the observed data with PC and NOTEARS most of the edges are not oriented(i.e direction) which is essential for treatment effect analysis. 

\begin{figure*}[htbp]
\floatconts
  {fig:no_tears}
  {\caption{Causal DAG generated with NoTears, $max_{iter}=100$, $h_{tol}=1e-08$, $w_{threshold}=0.5$}}
  {\includegraphics[height=3cm,keepaspectratio]{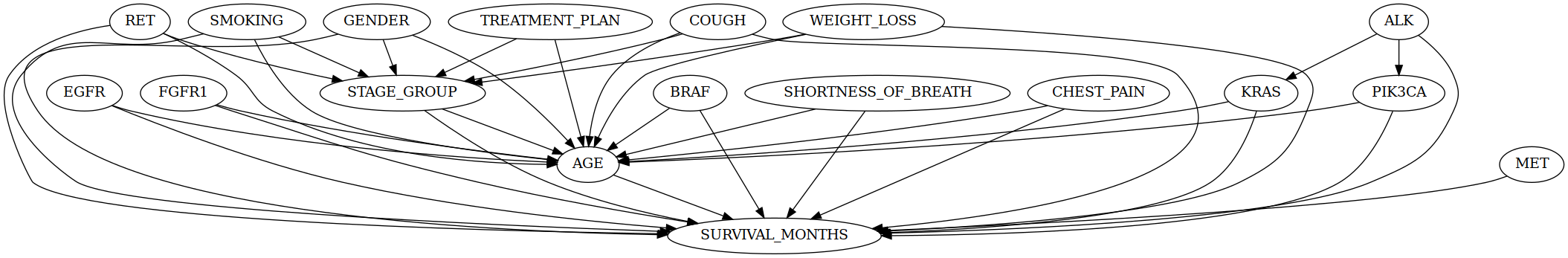}}
\end{figure*}

\begin{figure*}[htbp]
\floatconts
  {fig:PC}
  {\caption{DAG Generated using PC Algorithm}}
  {\includegraphics[height=6cm,keepaspectratio]{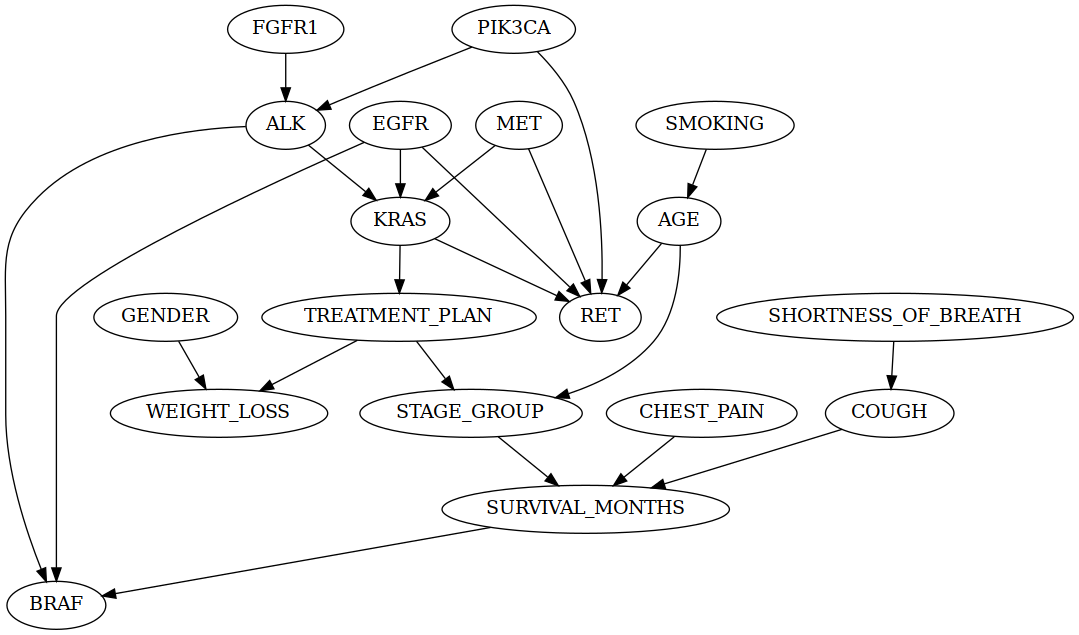}}
\end{figure*}

\begin{table*}[htbp]
\floatconts
  {tab:BDEU}
  {\caption{Bdeu score for the LLM Generated causal graph fitted to Bayesian Networks}}
  {
    \begin{tabular}{|c|c|c|c|c|c|}
    \hline
    Equivalent sample Size & Iterative Prompt & Single Prompt & Refined Single Prompt& NOTEARS &PC\\
    \hline
    5 & -4688 & -4263 & -4228 & -7037 & -6418 \\
    10 & -4559 & -4203 & -4171 & -6935 & -6202 \\
    15 & -4496 & -4179 & -4150 & -6886 & -6092 \\
    \hline
    \end{tabular}
  }
\end{table*}

\begin{table*}[htbp]
\floatconts
  {tab:ATE}%
  {\caption{ATE with different treatment category and mutation evidence}}%
  {%
\begin{tabular}{|l|c|c|c|c|c|c|c|c|}
    \hline
    \text{Treatment Category} & \text{KRAS} & \text{EGFR} & \text{FGFR1} & \text{ALK} & \text{MET} & \text{PIK3CA} & \text{BRAF} & \text{RET} \\
    \hline
    \text{Chemotherapy} & $0.028753$ & $0.027040$ & $0.024604$ & $0.027979$ & $0.025711$ & $0.028071$ & $0.024134$ & $0.030823$ \\
    \hline
    \text{Targeted Therapy} & $0.020828$ & $0.021891$ & $0.015868$ & $0.018419$ & $0.016813$ & $0.019083$ & $0.015762$ & $0.023738$ \\
    \hline
    \text{Immunotherapy} & $0.007267$ & $0.003912$ & $0.005502$ & $0.006562$ & $0.005807$ & $0.006607$ & $0.005356$ & $0.004674$ \\
    \hline
    \end{tabular}
 }
\end{table*}

\subsection{ Validation }

To validate the generated LLMs against an observed dataset, we choose to model each DAG as a Bayesian Network and fit said network to our data set. This Bayesian approach allows us to make probabilistic statements about the causal effect \( \beta \) based on both our prior beliefs and the observed data. It naturally incorporates uncertainty and provides a richer understanding than a single point estimate. For each model, the adjacency matrix was used to create a Bayesian Network. The Conditional Probability Distributions (CPD) were fit for each node using the Bdeu score as a criteria. Creating and optimizing the Bayesian network along with calculating the final Bdeu score was done using the \emph{pgmpy} Python package \cite{ankan2015pgmpy}.

For each of the Bayesian networks generated, the Bdeu score was calculated using the observational NSCLC dataset to compare how each model fit to the observed data. The score for the five models are given in Table \ref{tab:BDEU}. We note that the LLM methods vastly outperform optimization based approaches. The best score achieved through LLMs were $-4150$ while the NOTEARS and PC algorithm was only about to achieve $-6886$ and $-6092$ respectively. 

\subsection{Interpreting}
The Average Treatment Effect (ATE) \cite{10.1093/aje/kwad012} is a metric used in causal inference to quantify the variation in mean outcomes between treated and control group. We take our best fitting model, the single prompt LLM with updates, run causal inferencing methods to see the effect different biomarkers have of the treatment outcome. For this we calculate the ATE using the variable elimination method to compute the conditional probability of survival based on getting the treatment and having the mutation. 
\begin{equation}
    ATE = E[Y_{1} - Y_{0}]
    \label{eq:ate}
\end{equation}

where E is the expectation, $Y_1$ is treated and $Y_0$ is controlled group.
The ATE values for various treatments depending on certain gene mutations are shown in the Table \ref{tab:ATE}. The ATE on patients with a specific gene mutation compared to people without it. Different gene mutations are shown by KRAS, EGFR, and ALK columns, while chemotherapy and immunotherapy are indicated by rows. The information aids in understanding each treatment's efficacy for people with a particular mutation and directs individualised medical choices. We observe that the presence of a RET mutation has the largest effect for both Chemotherapy and Targeted Therapies. In the targeted therapy category the FGFR1 biomarker displaying the least interaction at $0.015868$. Notably, the immunotherapy treatment category had poor scores across all biomarkers. This is not too surprising as the most effective markers for Immunotherapy are PD-L1 expression and Tumor Mutational burden, neither of which are represented in the current set of markers \cite{yarchoan_pd-l1_nodate}. These results suggest varying degrees of influence of these treatment modalities on different biomarkers.  

\section{Conclusion \& Future Work}

As the true potential of LLMs have yet to be discovered, it is clear that there is potential for them to revolutionize healthcare \cite{karabacak_embracing_nodate}. In this article we have investigated how LLMs could be applied in one particular aspect of healthcare analysis, automating the generation of Causal DAGs. We investigated two different methods of DAG generation, an iterative query for each node and querying all individual nodes at once. We also looked at including human intervention(from in house pathological doctor) to condition on different nodes. Using the Bdeu score as a metric, we evaluated that LLM methods outperformed existing Causal Discovery methods.

In addition, further data modalities and features should be included to increase better characterize a patient's response to therapy. Currently we are using only 18 features for the analysis. The intricacy of cause effect analysis for NSCLC may not be fully captured by this restricted dataset, which could lead to an oversimplification of complicated causal linkages. Due to the fact that the data was extracted in 2018, the range of genes examined may not reflect all genes clinically relevant to NSCLC. Future work would expand the amount of genomic variations being examined but also include higher level genomic markers such as Tumor Mutational Burder (TMB) and Microsatellite Instability status (MSI) \cite{sha_tumor_2020}. In additional, other histopathological biomarkers, such as PD-L1, are key in diagnosing and determining treatment course \cite{bulutay_importance_2021}. Incorporating these additional features could bring greater insight into a patient's response to a given therapy.

The final limitation we observed is in our use of general purpose LLM models. While these LLMs have been trained on a wide variety of tasks, they lack the specialized knowledge that is needed to determine causal relationships in the medical domain. Future work would include looking at models that have been specially trained on medical literature such as Med-PaLM or BioGPT \cite{singhal_large_2023,luo_biogpt_2022}. In addition, specialized knowledge bases such as KEGG could be included as another source of truth for molecular associations \cite{kanehisa_kegg_2023}.

\clearpage





\bibliography{jmlr-sample}






\end{document}